%% file: obm_iros_main.tex
\documentclass[letterpaper, 10 pt, conference]{ieeeconf}  %

\IEEEoverridecommandlockouts                              %

\usepackage[backend=bibtex,
            hyperref=false,
            url=true,
            isbn=false,
            doi=false,
            backref=false,
            style=numeric-comp,
            sorting=none,
            block=none]{biblatex}

\renewcommand{\bibfont}{\small}
\input{packages}
\input{math_commands}

\input{macros}

\title{\LARGE \bf
Learning Object-Based State Estimators for Household Robots
}

\author{Yilun Du$^{1}$ \and Tomas Lozano-Perez$^{1}$ \and Leslie Pack Kaelbling$^{1}$%
\thanks{$^{1}$Computer Science and Artificial Intelligence Laboratory, MIT, USA, {\tt\small yilundu, tlp, lpk@mit.edu}.}}

\begin{document}

\maketitle

\thispagestyle{empty}
\pagestyle{empty}

\input{text/abstract}

\input{text/intro}

\input{text/related_work}

\input{text/method}

\input{text/evaluation}
\input{text/discussion}

\renewcommand*{\bibfont}{\footnotesize}
\begin{flushright}
\printbibliography %
\end{flushright}

\end{document}

%% file: packages.tex
\let\labelindent\relax
\usepackage{multirow}

\usepackage{algorithm, algorithmic}

\usepackage{color,xcolor}
\usepackage{epsfig}
\usepackage{graphicx}

\usepackage{adjustbox}
\usepackage{array}
\usepackage{booktabs}
\usepackage{colortbl}
\usepackage{float,wrapfig}
\usepackage{hhline}
\usepackage{multirow}
\usepackage{subcaption} %

\usepackage{bm}
\usepackage{nicefrac}
\usepackage{microtype}

\usepackage{changepage}
\usepackage{extramarks}
\usepackage{fancyhdr}
\usepackage{lastpage}
\usepackage{setspace}
\usepackage{soul}
\usepackage{xspace}

\usepackage{url}

\usepackage{enumerate}
\usepackage{todonotes} %

\usepackage{titlesec}
\usepackage{makecell}

\usepackage{mathtools}
\usepackage{stmaryrd}
\usepackage{hyperref}
\hypersetup{
    colorlinks=true,
    linkcolor=blue,
    filecolor=magenta,      
    urlcolor=cyan,
    }

%% file: math_commands.tex
\usepackage{amsmath,amsfonts,bm}

\def\eqref#1{equation~\ref{#1}}

\def\1{\bm{1}}

\def\vy{{\bm{y}}}
\def\vz{{\bm{z}}}

\DeclareMathAlphabet{\mathsfit}{\encodingdefault}{\sfdefault}{m}{sl}
\SetMathAlphabet{\mathsfit}{bold}{\encodingdefault}{\sfdefault}{bx}{n}

\def\sR{{\mathbb{R}}}

\DeclarePairedDelimiterX{\infdivx}[2]{(}{)}{%
  #1\;\delimsize|\delimsize|\;#2%
}

%% file: macros.tex
\newcolumntype{L}[1]{>{\raggedright\let\newline\\\arraybackslash\hspace{0pt}}m{#1}}
\newcolumntype{C}[1]{>{\centering\let\newline\\\arraybackslash\hspace{0pt}}m{#1}}
\newcolumntype{R}[1]{>{\raggedleft\let\newline\\\arraybackslash\hspace{0pt}}m{#1}}

\DeclareMathOperator*{\avg}{avg}

\usepackage{enumitem}
\setitemize{noitemsep,
topsep=0pt,parsep=0pt,partopsep=0pt,itemindent=0pt,leftmargin=20pt}
\setenumerate{noitemsep,
topsep=0pt,parsep=0pt,partopsep=0pt,itemindent=0pt,leftmargin=20pt}

\newcommand{\sect}[1]{Section~\ref{#1}}

\newcommand{\fig}[1]{Figure~\ref{#1}}
\newcommand{\tbl}[1]{Table~\ref{#1}}

\newcommand{\ignore}[1]{}
\newcommand{\norm}[1]{\lVert#1\rVert}

\makeatletter
\DeclareRobustCommand\onedot{\futurelet\@let@token\@onedot}
\def\@onedot{\ifx\@let@token.\else.\null\fi\xspace}

\def\eg{\emph{e.g}\onedot}

\makeatother

\definecolor{MyDarkBlue}{rgb}{0,0.08,1}
\definecolor{MyDarkGreen}{rgb}{0.02,0.6,0.02}
\definecolor{MyDarkRed}{rgb}{0.8,0.02,0.02}
\definecolor{MyDarkOrange}{rgb}{0.40,0.2,0.02}
\definecolor{MyPurple}{RGB}{111,0,255}
\definecolor{MyRed}{rgb}{1.0,0.0,0.0}
\definecolor{MyGold}{rgb}{0.75,0.6,0.12}
\definecolor{MyDarkgray}{rgb}{0.66, 0.66, 0.66}

\newcommand{\model}{OBM-Net\xspace}
\newcommand{\myparagraph}[1]{{\bf #1}\quad}

\newcommand{\cm}[1]{{\color{blue}#1}}

\newcommand{\daf}{{\sc daf}}
\newcommand{\mldaf}{{\sc ml-daf}}
\newcommand{\obm}{{\sc obm}}

%% file: text/abstract.tex
\begin{abstract}

A robot operating in a household makes observations of multiple objects as it moves around over the course of days or weeks.  The objects may be moved by inhabitants, but not completely at random.  The robot may be called upon later to retrieve objects and will need a long-term object-based memory in order to know how to find them.  Existing work in {\em semantic {\sc slam}} does not attempt to capture the dynamics of object movement. In this paper, we combine some aspects of classic techniques for data-association filtering with modern attention-based neural networks to construct object-based memory systems that operate on high-dimensional observations and hypotheses. We perform end-to-end learning on labeled observation trajectories to learn both the transition and observation models.  We demonstrate the system's effectiveness in maintaining memory of dynamically changing objects in both simulated environment and real images, and demonstrate improvements over classical structured approaches as well as unstructured neural approaches. Additional information available at project website: \href{https://yilundu.github.io/obm/}{https://yilundu.github.io/obm/}.
\end{abstract}

%% file: text/intro.tex
\section{Introduction}

Consider a robot operating in a household, making observations of multiple objects as it moves around over the course of days or weeks.  The objects may be moved by the inhabitants, even when the robot is not observing them, and we expect the robot to be able to find any of the objects when requested. We will call this type of problem {\em entity monitoring}. It occurs in many applications, but we are particularly motivated by robotics applications where the observations are very high dimensional, such as images or point clouds. 

One version of this problem is addressed by {\em semantic {\sc slam}}~\cite{Kostavelis} methods, which focus on constructing a metric map of the occupied space, keeping the robot localized, and detecting static objects and marking their locations in the map.  The SLAM aspects of these systems are excellent and we do not intend to replace or improve on them.  However, these methods generally do not model objects that might move (and in fact, actively work to avoid adding them to the map) and also do not fuse information about an object that might be acquired from multiple observations that are considerably spaced in time.  
Another version of this problem, {\em object tracking}, removes the SLAM considerations, and addresses situations in which observations are closely spaced in time and objects only briefly go out of view. 

This paper focuses instead on the problem of maintaining a memory of objects over a long time horizon, with multiple temporally and spatially distant observations of a single object and with objects that may be moved according to characteristic patterns over time. 
Solving such problems requires online {\em data association}, determining which individual objects generated each observation, and {\em state estimation}, aggregating the observations of each individual object to obtain a representation that is lower variance and more complete than any individual observation. This problem can be addressed by an online recursive {\em filtering} algorithm that receives a stream of object detections as input and generates, after each input observation, a set of hypotheses corresponding to the actual objects observed by the agent.

\input{new_fig_iros/teaser_right}

A classical solution to the entity monitoring problem, developed for the tracking case but extensible to other dynamic settings, is a {\em data association filter} (\daf) (the tutorial of~\cite{BarShalom09} provides a good introduction).  A Bayes-optimal solution to this problem can be formulated, but it requires representing a number of possible hypotheses that grows exponentially with the number of observations.   A much more practical, though less robust, approach is a maximum likelihood \daf{} (\mldaf), which commits, on each step, to a maximum likelihood data association:  the algorithm maintains a set of object hypotheses, one for each object (generally starting with the empty set) and for each observation it decides to either: (a) associate the observation with an existing object hypothesis and perform a Bayesian update on that hypothesis with the new data, (b) start a new object hypothesis based on this observation, or (c) discard the observation as noise.   As the number of entities in the domain and the time between observations of the same entity increase, the problem becomes more difficult and the system can begin to play the role of the long-term object-based memory (\obm) for an autonomous agent.

The engineering approach to constructing such an \obm{} requires many design choices, including the specification of a latent state-space for object hypotheses, a model relating observations to object states, another model specifying the evolution of object states over time, and thresholds or other decision rules for choosing, for a new observation, whether to associate it with an existing hypothesis, use it to start a new hypothesis, or discard it.  In any particular application, the engineer must tune all of these models and parameters to build an \obm{} that performs well.  This is a time-consuming process that must be repeated for each new application.

In this paper, we develop a method for training neural networks to perform as \obm{}s for dynamic entity monitoring.  \textit{In particular, we train a system to construct a memory of the objects in the environment, without explicit models of the robot's sensors, the types of objects to be encountered, or the patterns in which they might move in the environment.} Although it is possible to train an unstructured recurrent neural network (RNN) to solve this problem, we find that building in some aspects of the structure of the \obm\/ allows faster learning with less data and enables the system to address problems with a longer horizon.  We describe a neural-network architecture that uses self-attention as a mechanism for data association, and demonstrate its effectiveness in different robotics domains as illustrated in \fig{fig:teaser}. We first illustrate its application on a simulated robotic domain, enabling the estimation of position, type and shape of objects across time. We further validate our approach on real indoor images gathered in a domain with moving objects using a Jackal mobile robot.

%% file: new_fig_iros/teaser_right.tex
\begin{figure}
\centering
\includegraphics[width=\linewidth]{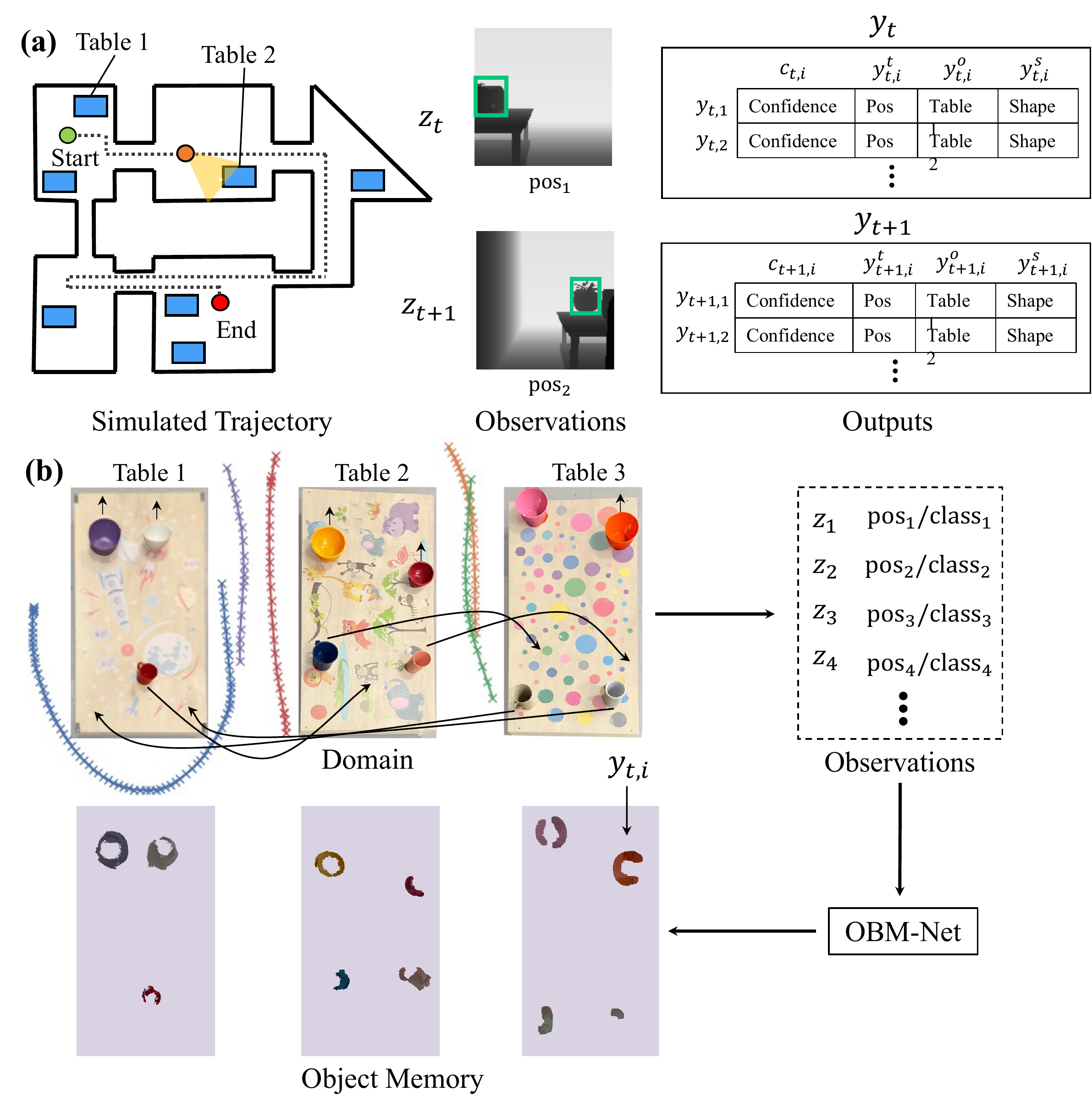}
\vspace{-15pt}
\caption{\small \textbf{(a) Simulated Domains.} \model takes a set of input observations generated from a trajectory exploring a house. At each timestep, \model receives an observation $\vz_t$, consisting of a single segmented depth map input of an object and its corresponding 2D offset on a table. At each timestep, \model outputs a set of predicted objects $\vy_t$ in the environment, where each object prediction $\vy_{ti}$ consists of a predicted confidence of the object and its associated position, shape, and table which it is on (objects move with dynamics across tables). \textbf{(b) Real Images.} \model takes a set of input observations made by a robot driving around three different tables at 5 separate points in time, where individual objects on tables migrate over time (indicated with arrows). Input observations are processed into a set of partial pointcloud detections, with each input observation $\vz_t$ to \model corresponding to the 2D position and class identity of a detected partial pointcloud. At each timestep, \model outputs a set of objects $\vy_t$, which when reconstructed correspond to predicted objects and their locations on each table.}
\label{fig:teaser}
\vspace{-20pt}
\end{figure}

%% file: text/related_work.tex
\section{Related Work}

\myparagraph{Semantic mapping}  Our work is related to a large literature in robotics on {\em semantic {\sc slam}}~\cite{chen2019suma++, zhang2018semantic, civera2011towards, cui2019sof, doherty2019multimodal, wang2019computationally} (Kostavelis et al.~\cite{Kostavelis} provide a good survey), which seeks to obtain a semantic labeling of places and objects in an environment. Typically, {\em semantic {\sc slam}} operates on static environments, with recent work on dynamic environments ~\cite{cui2019sof, wang2019computationally, yu2018ds, cui2020sdf} focusing on obtaining a semantic labeling of only the static places and objects in an environment. Similarly, Idrees et al.  ~\cite{idrees2020robomem, idrees2021were} explore the construction of a static object database in an environment. In contrast, in this paper, we explore a new setting, in which we are interested in capturing instead, a semantic labeling of the dynamically moving objects in a scene. Such a task is important for household robotics, where objects will be displaced by the occupants of the house.

\myparagraph{Learning Object Dynamics}  To operate a robot autonomously in a household environment, an underlying model of dynamics of objects in the house must be learned. Hawes et al.~\cite{hawes2017strands} constructs a architecture for such mobile robot autonomy, where a spatial-temporal clustering method~\cite{ambrus2015unsupervised} is used to extract dynamic objects from the environment.  Tipaldi et al.~\cite{tipaldi2013lifelong} represents the dynamics of objects by utilizing dynamic occupancy grids while Kucner et al.~\cite{kucner2013conditional} learning conditional probabilities of neighboring cells dynamic transitions in occupancy grid. Krajnik et al.~\cite{krajnik2017fremen} further explore model object dynamics utilizing periodic functions. In contrast, we propose to utilize a neural network in the form of \daf{} to construct a memory of objects in an environment, consisting of both associating objects across time, as well as constructing and learning their associated dynamics.

\myparagraph{Data Association}
Our work is inspired by past work on data-association filtering. The most classic filter, for the case of a single entity, is the Kalman filter~\cite{bishop2001introduction}.  In the presence of data-association uncertainty the Kalman filter can be extended by considering  assignments of observations to multiple existing hypotheses in a \daf{} or \mldaf{}.  
These approaches, all of which require hand-tuned transition and observation models, are described by~\cite{BarShalom09}. We show in supplementary results~\cite{obm_website} that our learned approach can learn the underlying transition and observation models and performs comparably to \mldaf{} with ground truth system dynamic and observation models on robotic domains.

A special case of the entity monitoring problem where observations are closely spaced in time has been extensively explored in the visual object tracking setting~\cite{luo2014multiple, xiang2015learning, Bewley2016_sort}.  In these problems, there is typically a fixed visual field populated with many smoothly moving objects.  This enables some specialized techniques that take advantage of the fact that the observations of each object are typically smoothly varying in space-time, and incorporate additional visual appearance cues. A related problem is {\em object re-identification}~\cite{girdhar2019cater, bai2019re, bansal2021did}, which is focused on matching images of specific people and cars, without taking any other information into account.

In contrast, in our setting, there is no fixed spatial field for observations, they may combine a variety of modalities, and may be temporally widely spaced, as would be the case when a robot moves through the rooms of a house, encountering and re-encountering different objects as it does so.

\myparagraph{Algorithmic priors for neural networks}
Our approach is an instance of a general technique that integrates algorithmic structure with end-to-end neural network training.  This approach has been applied to sequential decision making ~\cite{Tamar2016ValueIN}, particle filters ~\cite{Jonschkowski2018DifferentiablePF}, and Kalman filters ~\cite{Krishnan2015DeepKF}, as well as to a complex multi-module robot control system ~\cite{Karkus2019DifferentiableAN}.  The results generally are much more robust than completely hand-built models and much more sample-efficient than completely unstructured deep-learning.

%% file: text/method.tex
\newcommand{\inpspace}{\sR^{d_z}}
\newcommand{\outspace}{\sR^{d_y}}
\newcommand{\slotspace}{\sR^{d_s}}
\newcommand{\outweightspace}{(0, 1)}
\newcommand{\data}{\mathcal{D}}
\newcommand\ex[2]{#1^{(#2)}}
\newcommand{\loss}{\mathcal{L}}
\newcommand{\clusterloss}{\loss_{\rm obj}}
\newcommand{\slotloss}{\loss_{\rm slot}}
\newcommand{\sparseloss}{\loss_{\rm sparse}}

\setlength{\belowdisplayskip}{3pt} \setlength{\belowdisplayshortskip}{3pt}
\setlength{\abovedisplayskip}{3pt} \setlength{\abovedisplayshortskip}{3pt}

\section{Problem formulation}
\label{sect:loss}

We formalize the process of learning an object-based memory system (\obm{}).  When the \obm{} is executed online, it receives a stream of input observations $z_1, \ldots z_T$ where $z_t \in \inpspace$, and after each input $z_t$, it will output two vectors representing a set of predicted properties of hypothesized objects (\eg position, type, and shape) $y_t =[y_{tk}]_{k \in (1..K)}$ and an associated confidence score for each hypothesis, $c_t = [ c_{tk}]_{k \in (1..K)}$, where $y_{tk} \in \outspace$, $c_{tk} \in \outweightspace$. To ensure that confidences are bounded, we constrain $\sum_k c_{tk} = 1$.  We limit the maximum number of hypothesis ``slots'' in advance to $K$.
Dependent on the application, the $z$ and $y$ values may be in the same space with the same representation, but this is not necessary.

We have training data representing $N$ different entity-monitoring problem instances, 
\[\data = \{(\ex{z}{i}_t, \ex{m}{i}_t)_{t \in (1..L_i)} \}_{i \in (1..N)}\;\;,\]
where each training example is an input/output sequence of length $L_i$, each element of which consists of a pair of input $z$ and $m = \{m_j\}_{j \in (1..\ex{J}{i}_t)}$, which is a set of nominal object hypotheses representing the true {\em current state} of objects that have actually been observed so far in the sequence.  It will always be true that $\ex{m}{i}_t \subseteq \ex{m}{i}_{t+1}$ and $\ex{J}{i}_t \leq K$ because the set of objects seen so far is cumulative.

Our objective is to train a recurrent network to perform as an \obm{} effectively in problems that are drawn from the same distribution over latent domains as those in the training set. To do so, we formulate a network (described in section~\ref{sec:model}) with parameters $\theta$, which transduces the input sequence $z_1, \ldots, z_L$ into an output sequence $(y_1, c_1), \ldots, (y_L, c_L)$, 
and train it to minimize the following loss function:%
\begin{align*} \small
\mathcal{L}(\theta ; \data) = \sum_{i=1}^N \sum_{t = 1}^{L_i} 
    \clusterloss(\ex{y}{i}_t, \ex{m}{i}_t) &+
    \slotloss(\ex{y}{i}_t, \ex{c}{i}_t, \ex{m}{i}_t)  \\
    &+ \sparseloss(\ex{c}{i}_t) \;\;.%
\end{align*}
The $\clusterloss$ term is a {\em chamfer loss}, which looks for the predicted $y_k$ that is closest to each actual $m_j$ and sums their distances, making sure the network has found a good, high-confidence representation for each true object, with $\epsilon \ll 1$:%
\begin{equation*} \small
\clusterloss(y, \cm{c}, m) = \sum_j \min_k \frac{1}{c_k + \epsilon}\norm{y_k - m_j}\;\;.%
\end{equation*}
The $\slotloss$ term is similar, but makes sure that each object the network has found is a true object, where we multiply by $c_k$ to not penalize for predicted objects in which we have low confidence:
\begin{equation*} \small
\slotloss(y, c, m) = \sum_k \min_j c_k \norm{y_k - m_j}\;\;.%
\end{equation*}
Finally, the sparsity loss discourages the network from using multiple outputs to represent the same true object, by encouraging sparsity in object hypothesis confidences:%
\begin{equation*} \small
\sparseloss(c) = -\log \norm{c}\;\;.
\end{equation*}
Details in~\cite{obm_website} illustrate how such a loss encourages sparsity among attention weights.

\section{\model{}s}

\label{sec:model}
\label{sect:mem}
Inspired by the the basic form of classic \daf{} algorithms and the ability of modern neural-network techniques to learn complex networks, we have designed the \model{} architecture for learning \obm{}s and a customized procedure for training it from data, motivated by several design considerations.
First, because object hypotheses must be available after each individual input and because observations will generally be too large and the problem too difficult to solve from scratch each time, the network will have the structure of a recursive filter, with new memory values computed on each observation and then fed back for the next. Second, because the loss function is {\em set based}, that is, it doesn't matter what order the object hypotheses are delivered in, our memory structure should also be permutation invariant and independent of the number of objects, and so the memory processing is in the style of an attention mechanism.  Finally, in applications where the observations $z$ may be in a representation not well suited for hypothesis representation and aggregation, the memory operates on a latent representation that is related to observations and output hypotheses via encoder and decoder modules. 
 
 Figure~\ref{fig:arch} shows the architecture of the \model{} network.  The memory of the system is stored in $s$, which consists of $K$ elements, the $K$ hypotheses in \daf{}, each in $\slotspace$;  the length-$K$ vector $n$ of positive values encodes how many observations have been assigned to each slot during the execution so far.  New observations are combined with the memory state, and the state is updated to reflect the passage of time by a neural network constructed from seven modules with trainable weights.

 \input{figText/arch} 

  When an observation $z$ arrives, it is immediately {\bf encode}d into a vector $e$ in $\slotspace$, which is fed into subsequent modules.  First, {\bf attention} weights $w$ are computed for each hypothesis slot, using the encoded input and the existing content of that slot, representing the degree to which the current input ``matches'' the current  value of each hypothesis in memory, mirroring the hypothesis matching procedure in \daf{}s.  Since an observation typically matches only a limited number of hypotheses in \daf{}s, we force the network to commit to a sparse assignment of observations to object hypotheses while retaining the ability to effectively train with gradient descent, the {\bf suppress} module sets all but the top $M$ values in $w$ to 0 and renormalizes, to obtain the vector $a$ of $M$ values that sum to 1:
  \begin{equation*}\small
  w_k = \frac{\exp({\textbf{attend}(s_k, n_k, e)})}
              {\sum_{j=0}^n \exp({\textbf{attend}(s_j, n_k, e)})}\;\;
              ;\;\; a = \textbf{suppress}(w)\;\;.
 \end{equation*}
The $a$ vectors are integrated to obtain $n$, which is normalized to obtain the output confidence $c$.

Mirroring hypothesis updates in \daf{}s, the {\bf update} module also operates on the encoded input and the contents of each hypothesis slot, producing a hypothetical update of the hypothesis in that slot under the assumption that the current $z$ is an observation of the object represented by that slot;  so 
 for all slots $k$, \[u_k = \textbf{update}(s_k, n_k, e)\;\;.\]  

To enable the rejection of outlier observations, a scalar {\bf relevance} value, $r \in (0, 1)$, is computed from $s$ and $e$;  this value modulates the degree to which slot values are updated, and gives the machine the ability to ignore or downweight an input.  It is computed as
\begin{equation*} \small
 r = \textbf{relevance}(e, s, n) ={\rm NN}_2(\avg_{k=1}^K{\rm NN}_1(e, s_k, n_k))\;\;,  
\end{equation*}
where ${\rm NN}_1$ is a fully connected network with the same input and output dimensions and ${\rm NN}_2$ is a fully connected network with a single sigmoid output unit.
The attention output $a$ and relevance $r$ are now used to decide how to combine all possible slot-updates $u$ with the old slot values $s_t$ using the following fixed formula for each slot $k$:
\begin{equation*} \small
    {s'_{t}}_k = (1 - r a_k) {s_t}_k + r a_k u_k\;\;.
\end{equation*}
Because most of the $a_k$ values have been set to 0, this results in a sparse update which will ideally concentrate on a single slot to which this observation is being ``assigned''.

To obtain outputs, slot values $s'_{t}$ are then {\bf decoded} into the outputs, $y$, using a fully connected network:%
\begin{equation*} \small
    y_k = \textbf{decode}({s'_{t}}_k)\;\;.    
\end{equation*}%
Finally, to simulate transition updates in \daf{}s and to handle the setting in which object state evolves over time, we add a {\bf transition} module, which computes the state $s_{t+1}$ from the new slot values $s_{t}'$ using an additional neural network: 
\begin{equation*} \small
    {s_{t+1}}_k = \textbf{transition}(s'_{t})_k\;\;.
\end{equation*}
These values are then fed back, recurrently, as inputs to the overall system.

Given a data set $\data$, we train the \model{} network end-to-end to minimize loss function $\loss$, with a slight modification.  We find that including the $\sparseloss$ term from the beginning of training results in poor learning, but adopting a training scheme in which the $\sparseloss$ is first omitted then reintroduced over training epochs, results in reliable training that is efficient in both time and data.

%% file: figText/arch.tex
\begin{figure*}
\centering

\begin{minipage}{0.6\textwidth}
\centering
\includegraphics[width=\linewidth]{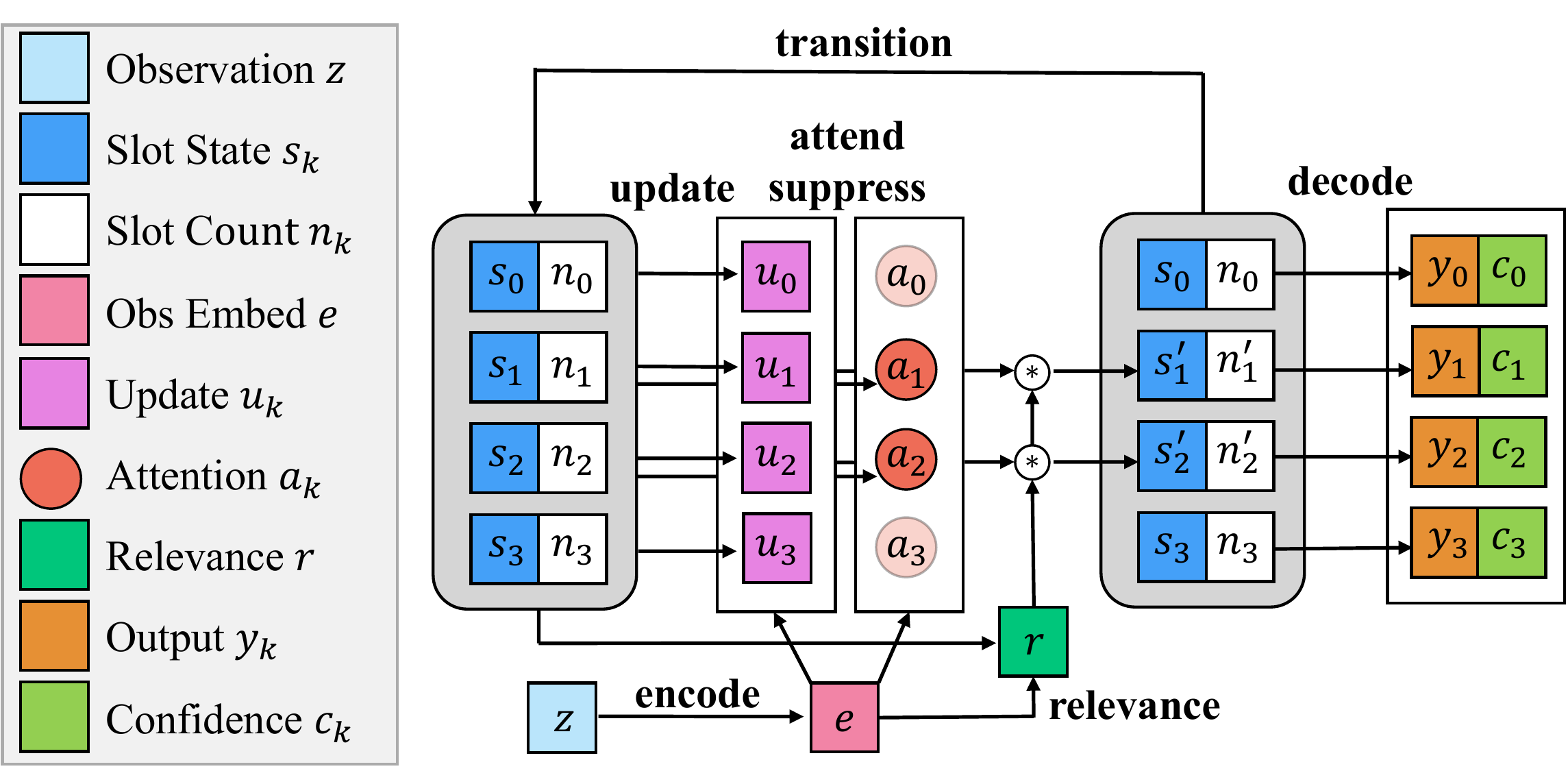}

 \end{minipage} 
 \scalebox{0.78}{
 \begin{minipage}{0.40\textwidth}
\vspace{-20pt}
\begin{algorithm}[H]
\small
\begin{algorithmic}
    \STATE \textbf{Input:} input observations $z_1, \ldots, z_T$, count $n \in \mathbb{R}^K$, state $s \in \mathbb{R}^{K \times D}$ 
    \FOR{timestep $t = 1$ to $T$}
    \STATE $e \gets \textbf{encode}(z_t)$
    \STATE $r \gets \textbf{relevance}(e, s, n)$
    \FOR{slot $k = 1$ to $K$}
    \STATE $a_k \gets \textbf{suppress}(\textbf{attend}(s_k, n_k, e))$
    \STATE $u_k \gets \textbf{update}(s_k, n_k, e)$
    \STATE $s'_k \gets (1 - r a_k) s_k + r a_k u_k$
    \STATE $n_k \gets n_k + a_k$
    \STATE $y_k \gets \textbf{decode}(s'_k)$
    \STATE $c_k \gets n_k / (\sum_i n_i)$
    \STATE $s_k \gets \textbf{transition}(s'_k)$
    \ENDFOR
    \ENDFOR
  \end{algorithmic}

 \caption{\small \model execution.}
 \end{algorithm}
 \end{minipage}
 }

\vspace{-5pt}
 \caption{\small Architecture and pseudocode of \model{}. Observations are fed  sequentially to \model{}, and encoded with respect to each hypothesis. A subset of the hypotheses are updated at each time-step, with corresponding slot counts incremented according to attention weight. Slots are then decoded, with the confidence of an output proportional to underlying slot count.} 
\label{fig:arch}
\vspace{-15pt}
\end{figure*}

%% file: text/evaluation.tex
\section{Empirical results}
\label{sect:emp}

We evaluate \model on several different {\em entity monitoring} tasks.  Additional information in ~\cite{obm_website} provides results in a large number of simple, illustrative tasks, illustrating the ability of \model to learn to perform as a data-association filter in a variety of situations, some with observations in a very different space from than the desired output hypotheses.  In this paper, we focus on dynamic entities in robotics domains.
First, we evaluate the performance of \model on the complex simulated household robot domain shown in  \fig{fig:teaser}(a), and then validate the ability of \model to capture an object with underlying dynamics and complex properties, as well as its utility for downstream robotics object-fetching tasks. Next, we evaluate the performance of \model on real images captured by a mobile robot in \fig{fig:teaser}(b).

\subsection{Simulated Household Robot Domains}
\label{sect:simulation}

\input{hybrid_figText/robot_3d_reconstruct}

We first validate that \model can learn to solve the entity monitoring task in simulated robotic settings. 

\myparagraph{Setup.} We model a robot moving within a house, as pictured in \fig{fig:teaser}(a), in the PyBullet simulation environment. In this house, each problem will involve following a trajectory consisting of a sequence of 50 locations.  These locations are distributed across 5-6 separate rooms, with later time steps potentially revisiting earlier locations.  At each location, the robot looks around and if there is a table within view (which happens about 50\% of the time), it will get an observation of one of the objects on the table or an empty observation otherwise.  Each new problem has 8 tables whose locations are drawn from a larger set of potential table locations and on each table there will be two objects drawn from a small set of classes, e.g. lamp, cushion, etc.  Each object class has a characteristic stochastic movement pattern, with one object class sequentially teleporting between tables (details in ~\cite{obm_website}).  The goal is for the robot to be able to construct hypotheses for each distinct object it has seen and to be able to predict for each object the table it is currently on and its location relative to the table. 
~
More precisely, the input sequence of observations $z$ corresponds to a segmented depth map of a single object visible given the camera pose at a particular location in the trajectory (or an empty observation in the case no object is visible), as well as which table it is resting on and its positional offset relative to the table. The desired output $y$ values are, for each distinct object seen so far, the predicted table $y^t$ it is on currently as well as its associated offset relative to the predicted table, $y^o$.

We train on a total of 10000 randomly sampled trajectories in the same floor plan, but with new randomly drawn object instances and tables for each trajectory. We test using 1000 trajectories,  with test object meshes drawn from a set {\em disjoint from} the set of object meshes used during training (but sharing the same semantic classes). To test the flexibility of the approach, we consider three different configurations of object classes on tables (each object class contains roughly 20 meshes, with 50\% of meshes used for training and the other 50\% used for testing).

\input{new_fig_iros/configuration}
\input{figText/pr2}

\begin{itemize}
    \item Configuration A: Plants (which move horizontally across a table), cushions (which move vertically across a table), and baskets (which move diagonally in a table, but also teleport sequentially to different tables).
    \item Configuration B: Table Lamps (which move horizontally), Trash Cans (which move vertically) and cushions (which move diagonally across a table, but also sequentially teleport between tables).
    \item Configuration C: Cushions (which move horizontally), Table Lamps (which move vertically) and plants (which move diagonally across a table, but also sequentially teleport between tables).
\end{itemize}
We illustrate the underlying dynamics of objects in each configuration in \fig{fig:configuration}.

{\em Crucially, none of the category shape information or dynamics is built in:  the observation and transition models as well as the desired output representation are learned entirely from data.}

\myparagraph{Metrics.} To test the efficacy of our approach, we measure to what extent each hypothesis slot $m_i$ can recover both the table that the associated object is on, as well as the object's position relative to the table. We match a hypothesis slot $k$ with each object label $y_i$ by computing $\textrm{arg} \min_k \norm{y_i^o - m_k^o} + \text{Loss}_{\text{CE}}(y_i^t - m_k^t)$. For each match, we report the accuracy of $m_k^t$ matching $y_i^t$, and as well the mean absolute error between $y_i^o$ and $m_k^o$. When the table prediction for $y_i$ is incorrect, we set mean absolute error to be equal to half the table size (0.15), as reported table offsets are  meaningless in that case. In this setting, both \model and associated baselines use 10 hypothesis slots.

\myparagraph{Baselines.} We first compare \model to online learned baselines of LSTM~\cite{Hochreiter1997Long} and Set Transformer~\cite{lee2018set} where all learned network architectures are structured to use $\sim 50000$ parameters. We further compare to task-specific baselines for this problem (detail at ~\cite{obm_website}). First, given a localized input-segmented depth map, we extract object offsets by averaging all points in the point cloud associated with each segment. To associate objects dynamically across time, we use batch K-means clustering on the inferred object candidate offsets and  associated table identities to obtain a set of objects.  We further compare \model with the more complex spatial-temporal clustering method used in the STRANDS project~\cite{hawes2017strands} to infer objects in a real robotic setup from our underlying segmented depth maps, as well as a hand-crafted DAF system using ground truth dynamics. Finally, we compare our approach with {\em semantic {\sc slam}} approach, DS-SLAM~\cite{yu2018ds} for segmenting dynamically changing objects. For all learned models, we convert the segmented depth maps into downsampled 3D pointclouds.

\myparagraph{Results.} \tbl{tbl:comparison_robotics} shows that \model outperforms the baselines in both estimating the supporting tables and regressing the relative position of the objects across different numbers of observations. \fig{fig:object_recovery} shows the prediction error of all methods as a function of the number of steps since the robot last saw an object; observe that \model is substantially better at long-term memory than the LSTM and Set Transformer, and still outperforms the clustering and STRANDS baselines even with long inter-observation gaps. As an upper bound, we compare with an oracle model, which knows ground truth object identity and dynamics (ignoring object collision). We find that \model performs similarly to the oracle model (performance across all models drops due to stochasticity), and in some cases does better, perhaps by modeling object collisions (not modeled in the oracle model).

\myparagraph{Shape reconstruction.} By adding a shape occupancy prediction head~\cite{mescheder2019occupancy} to \model, we can also regress the underlying 3D shapes of our objects. We predict each shape at $32 \times 32 \times 32$ resolution, decoding each occupancy at each voxel coordinate using a MLP head conditioned on a hypothesis state. Quantitatively, we find that our approach gets 95.33\% accuracy compared to 72.74\% accuracy obtained by a LSTM and 73.67\% obtained by a Set Transformer when predicting voxels for each test mesh in the test set. We provide visualization of predicted shapes from \model in \fig{fig:3d_vis}.

\input{figText/3d_vis}

\input{figText/object_movement}
\input{new_fig_iros/data_preprocess}

\myparagraph{Downstream Tasks.} Finally, we verify that object hypotheses from \model can usefully support a task in which a robot has to retrieve an object it has previously observed. First, we consider the task of finding a previously-encountered object. We train LSTM, Set Transformer, and \model to predict underlying object class $y^{c}$ for each object hypothesis, as well as shape estimate and location. Given a desired object class (for example, either a plant, cushion, or bucket in configuration A) we wish to find, the robot examines each prediction $(y_i, c_i)$ and navigates in the simulated world to look for an object of the specified class, based on predictions of $y_i^{t}$ and $y_i^{o}$.   It first goes to the most confident location of an object of that class, then if it does not locate the object there, it goes to the next most confident, and so on.  We measure the number of predictions that need to be queried to find the object, as well as the percentage of trials in which the robot succeeded within 10 attempts. On this task, we find that a LSTM obtains an overall planning success of 68.75\% with an average number of 5.38 hypotheses investigated before finding an object. In contrast, the Set Transformer obtains a planning success of 81.25\% with on average 4.88 attempts. We find that \model performs best and is able to find the object of the desired class  100\%  of the time, with an average of 2.03 hypotheses examined before finding the object. 

Next, we qualitatively analyze the 3D reconstructions of each object hypothesis with respect to its ability to support long-horizon manipulation planning.  Although final execution of the grasp of an object can often be done based on a partial point-cloud from a single view, it is useful to be able to make predictions about the 3D reconstruction.  These can be used for shape-completion of the current view as well as for planning an overall approach to manipulating an object before the object is in direct view of the robot.  To test the functional utility of the 3D reconstructions, we compute grasps on the underlying shape by looking for parallel planar surfaces large enough to accommodate the gripper.  We then try to execute that grasp on the target 3D object we wish to grasp in the (simulated) real world. As illustrated in \fig{fig:object_grasp}, we find that the 3D reconstruction of object hypotheses from \model is accurate enough to enable grasping of a real 3D shape. In contrast, predictions from LSTM and Set Transformer baselines are significantly poorer and do not enable downstream manipulation.

\subsection{Real Robot Domain}
\label{sect:real_robot}

\input{new_fig_iros/3d_reconstruct_real_side}

We next validate that \model can solve the dynamic entity monitoring task on real RGB-D video captured by a LiDAR Camera L515 mounted on a mobile robot. Please see the attached supplemental video for captured footage of our approach as it is running, as well as reconstructed objects obtained by \model.

\myparagraph{Setup.} We capture images from the robot moving between three separate tables across five separate points in time.  Each observation point consists of a trajectory, during which local observation information is fused.  At the end of each trajectory, called a time-step in the following, a set of observations are presented to the \model.

We illustrate the trajectories and placement of tables at each time-step in \fig{fig:teaser}(b) (blue = time step 1, red = time step 2, green = time step 3, yellow = time step 4, purple = time step 5), with robot looking at table 1 in time steps 1 and 5, table 2 on time steps 2 and 4, and table 3 on time step 3. At each time step, bowls are displaced vertically on their table, while mugs are moved between adjacent tables (illustrated in \fig{fig:object_movement}). In this setting, the input sequence of observations $z$ to \model correspond to the 2D offset of a detected object with respect to the center of the table it lies on, the table it lies on and its associated type.  The desired output $y$ values are, for each object seen so far, the predicted table $y^t$ it is on currently as well as its associated offset relative to the predicted table, $y^o$. For computational efficiency, we use existing local geometric alignment algorithms to reconstruct object shapes from predicted offsets $y^o$, as opposed to directly regressing shape from \model as done in \sect{sect:simulation}.  We made this choice to enable efficient execution of our system on a robot, but it is straightforward to directly train \model to reconstruct shapes as done in \sect{sect:simulation}.

To extract object detections at each time step, we first compute camera poses for each captured image in the sequence using ORBSLAM2~\cite{mur2017orb}. Next, we extract partial point clouds of observed mugs and bowls on the table from each image by combining Mask-RCNN~\cite{he2017mask} segmentations of each RGB image with its corresponding depth image (illustrated in \fig{fig:data_preprocess}). Finally, we fuse partial point clouds across images in a time step using the camera poses from ORBSLAM2 to obtain a set of object detections at that time step. We then compute the 2D offset of a detected object by computing the centroid of the partial point cloud. We train \model on 1000 simulated trajectories of objects following the same movement dynamics and measure the ability of \model to generalize to real data. We quantitatively measure object accuracy, the percentage of objects correctly inferred at each table at time step 5, as well as position error, the distance of the nearest inferred object's position from the closest actual object's position on a table.

\input{new_fig_iros/real_robot_eval}
\input{new_fig_iros/3d_reconstruct_real_top}

\myparagraph{Baselines.} We compare \model to two of the best-performing baseline approaches in \sect{sect:simulation} to obtain a semantic segmentation of the dynamically changing mugs and bowls on real images. First, we consider the clustering baseline from \sect{sect:simulation}, where we directly aggregate underlying semantic detections across time and apply clustering to the resultant outputs, using the ground truth number of objects as the number of centers. Next, we adopt the semantic SLAM approach of~\cite{yu2018ds} to deal with dynamic objects, where semantic segmentations of objects are updated based on new observations if they meet a log-odds probability threshold.  Neither baseline has a model of dynamics of the objects.

\myparagraph{Results.} We illustrate the reconstructions of predicted objects on each table at time step 5 from \model and each of the baselines, using a side view in \fig{fig:3d_reconstruct_side} and from a vertical view in \fig{fig:3d_reconstruct_top}.  In \fig{fig:3d_reconstruct_side} reconstructions of individual shapes from \model can be seen to be more complete compared to those of the baselines. \model is also able to aggregate information about a particular object seen from past observations of the object at the same and other tables. In \fig{fig:3d_reconstruct_top}, reconstructions of individual shapes can be seen to satisfy the underlying dynamics of objects, with horizontal movement of bowls on each table, as well as the movement of mugs across tables. In particular, in both \fig{fig:3d_reconstruct_side} and \fig{fig:3d_reconstruct_top}, while \model is able to reconstruct the correct mug object at table 2 and 3, both DS-SLAM and clustering baselines reconstruct the incorrect mug objects. Furthermore, on table 2, \model can reconstruct pink and blue mugs more accurately by aggregating views of the mugs seen at other tables.  We further quantitatively evaluate our approach compared to baselines in \tbl{tbl:comparison_robotics_real} and find that \model substantially outperforms the baselines.

%% file: hybrid_figText/robot_3d_reconstruct.tex
\begin{table*}[t]
\footnotesize
\setlength{\tabcolsep}{3.5pt}
\centering
\resizebox{\linewidth}{!}{
\begin{tabular}{lcccccccccccccccccc}
    \toprule
    & \multicolumn{6}{c}{Configuration A} & \multicolumn{6}{c}{Configuration B} & \multicolumn{6}{c}{Configuration C} \\
     \cmidrule(lr){2-7} \cmidrule(lr){8-13} \cmidrule(lr){14-19}
      & \multicolumn{3}{c}{Table Accuracy}  & \multicolumn{3}{c}{Position Error} &  \multicolumn{3}{c}{Table Accuracy}  & \multicolumn{3}{c}{Position Error} &  \multicolumn{3}{c}{Table Accuracy}  & \multicolumn{3}{c}{Position Error} \\
    
     \cmidrule(lr){2-4} \cmidrule(lr){5-7} \cmidrule(lr){8-10} \cmidrule(lr){11-13} \cmidrule(lr){14-16} \cmidrule(lr){17-19}
      Observations & 10 & 25 & 50  & 10 & 25 & 50 &  10 & 25 & 50 & 10 & 25 & 50 &  10 & 25 & 50  & 10 & 25 & 50 \\
     \midrule
      \textbf{Non Learned} \\
      \quad Clustering  &  0.761 & 0.695 & 0.485 & 0.053 & 0.070 & 0.103 & 0.761 & 0.695 & 0.488 & 0.053 & 0.070 & 0.103 & 0.761 & 0.695 & 0.488 & 0.053 & 0.069  & 0.103   \\
      \quad STRANDS &  0.900 & 0.733 & 0.610 & 0.033 & 0.057 & 0.085 &  0.940 & 0.841 & 0.737 & 0.023 & 0.048 & 0.087 &  0.973 & 0.832 & 0.774 & 0.031 & 0.055 & 0.086  \\
      \quad DS-SLAM~\cite{yu2018ds}  & 0.953 & 0.769 & 0.641 & 0.020 & 0.045 & 0.080  & 0.924 & 0.858 & 0.775 & 0.021 & 0.047 & 0.083 & 0.971 & 0.898 & 0.776 & 0.032 & 0.053 & 0.083 \\
      \quad DAF &  0.959 & 0.807 & 0.670 & 0.022 & 0.043 & 0.081 &  0.937 & 0.871 & 0.787 & 0.021 & 0.047 & 0.084 &  0.974 & 0.914 & 0.803 & 0.030 & 0.053 & 0.083  \\
     \textbf{Learned} \\
      \quad Set Transformer  &  0.883 & 0.619 & 0.476 & 0.034 & 0.066 & 0.089 & 0.919 & 0.771 & 0.542 & 0.024 & 0.052 & 0.093  & 0.885 & 0.745 & 0.649 & 0.037 & 0.056 & 0.089 \\
      \quad LSTM  &  0.839 & 0.661 & 0.406 & 0.058 & 0.093 & 0.126 & 0.875 & 0.716 & 0.514 & 0.053 & 0.094 & 0.123 & 0.892 & 0.717 & 0.519 & 0.052 & 0.091 & 0.130  \\
       \quad \model&  \textbf{0.984} & \textbf{0.926} & \textbf{0.809} &  \textbf{0.019} & \textbf{0.041} & \textbf{0.078} & \textbf{0.989} & \textbf{0.924} & \textbf{0.795} & \textbf{0.021} & \textbf{0.046} & \textbf{0.082}  &  \textbf{0.988} & \textbf{0.932} & \textbf{0.873} & \textbf{0.027} & \textbf{0.052} & \textbf{0.080} \\
     \bottomrule

\end{tabular}
}
\caption{\small \textbf{Quantitative Analysis of \model on Simulated Household Domain.} Quantitative comparison of \model with baselines across 3 studied household domain configurations across 10, 25, 50 observations. 
}
\label{tbl:comparison_robotics}
\vspace{-15pt}
\end{table*}

%% file: new_fig_iros/configuration.tex
\begin{figure}
\centering
\includegraphics[width=1.0\linewidth]{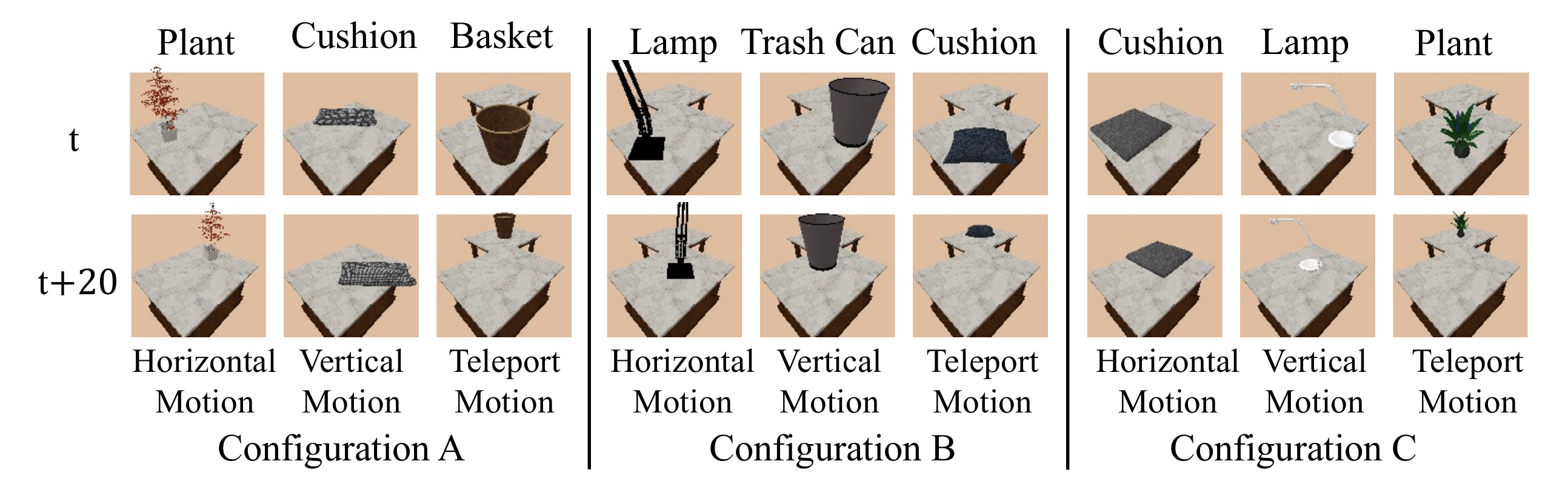}
\caption{\small \textbf{Simulated Object Dynamics.} Illustration of object dynamics in simulation across each class of objects in each of the 3 configurations. Images illustrate 20 timesteps of motion.} 
\label{fig:configuration}
\vspace{-15pt}
\end{figure}

%% file: figText/pr2.tex
\begin{figure}[t]
    \centering
    \includegraphics[width=1.0\linewidth]{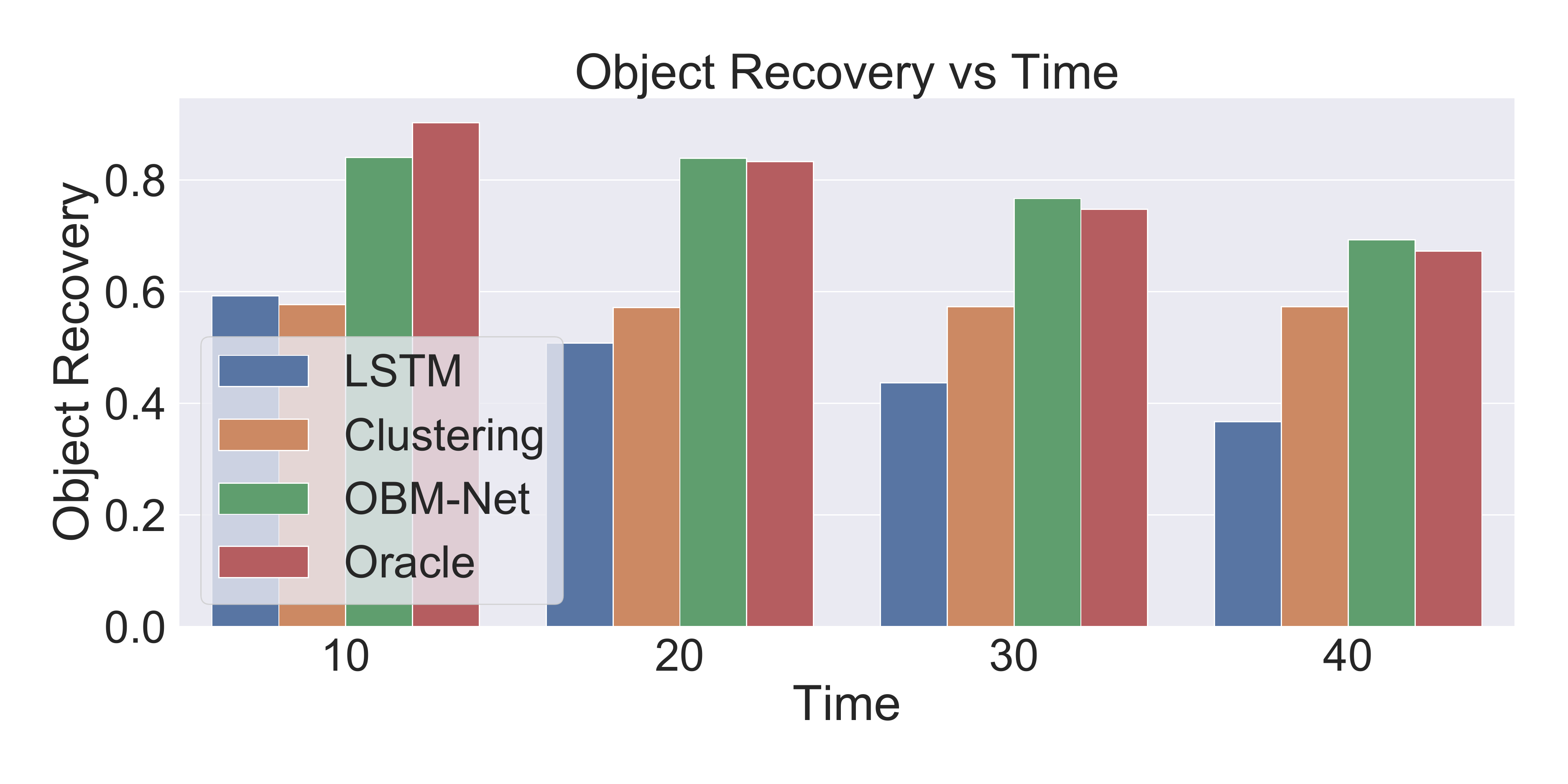}
    \vspace{-25pt}
    \caption{\small \textbf{Object Recovery over Time.}  Percentage of objects correctly recovered as a function of timesteps since seeing the object last. \model performs similarly to an oracle with ground truth dynamics. }
    \label{fig:object_recovery}
    \vspace{-20pt}
\end{figure}

%% file: figText/3d_vis.tex
\begin{figure}[t]
\centering
\includegraphics[width=0.9\linewidth]{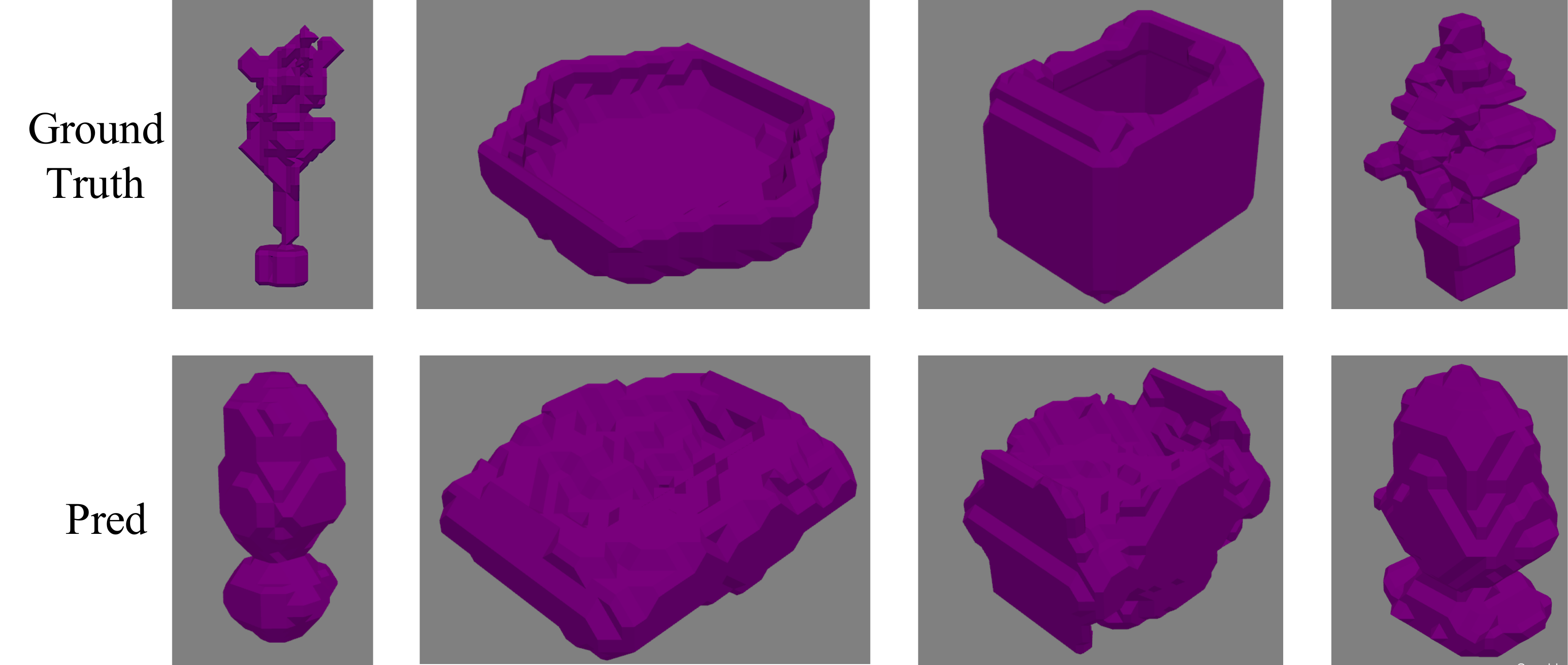}
\caption{\small \textbf{3D Shape Reconstruction.} Illustration of predicted 3D shape reconstructions using \model on unseen shapes at test time, compared to corresponding ground truth shape. \model is able to capture the coarse detail of individual shapes. }
\label{fig:3d_vis}
\vspace{-5pt}
\end{figure}

\begin{figure}[t]
    \centering
    \includegraphics[width=1.0\linewidth]{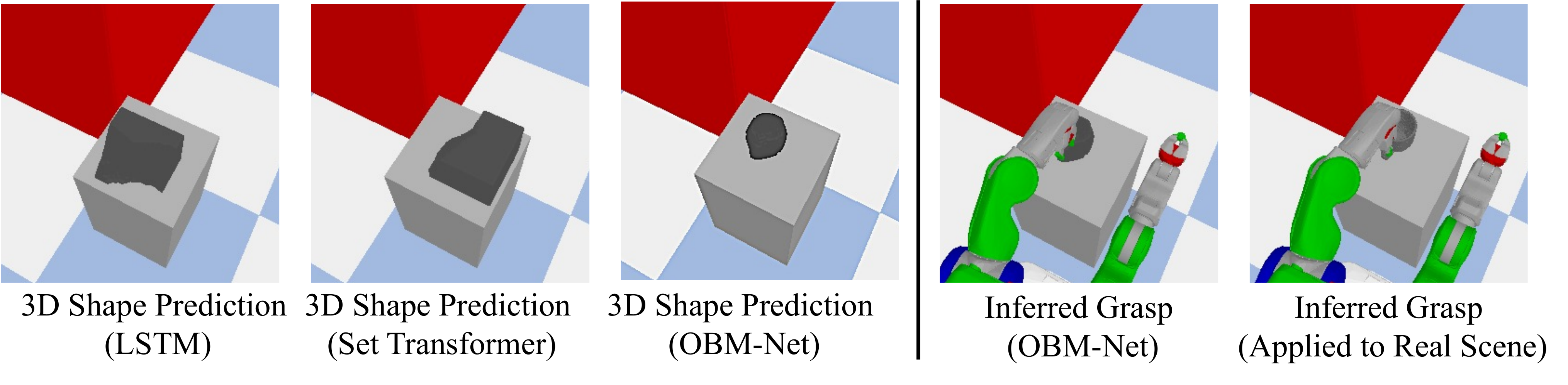}
    \vspace{-15pt}
    \caption{\small \textbf{(Left) 3D Reconstructions.} Illustration of 3D reconstructions of hypothesis from each model. \model obtains accurate 3D reconstructions. \textbf{(Right) Estimated Grasps.} We utilize the predicted 3D mesh from \model to infer a grasp which successfully enables the grasp of a real object in the ground truth scene.  }
    \label{fig:object_grasp}
    \vspace{-20pt}
\end{figure}

%% file: figText/object_movement.tex
\begin{figure}
\centering
\includegraphics[width=0.9\linewidth]{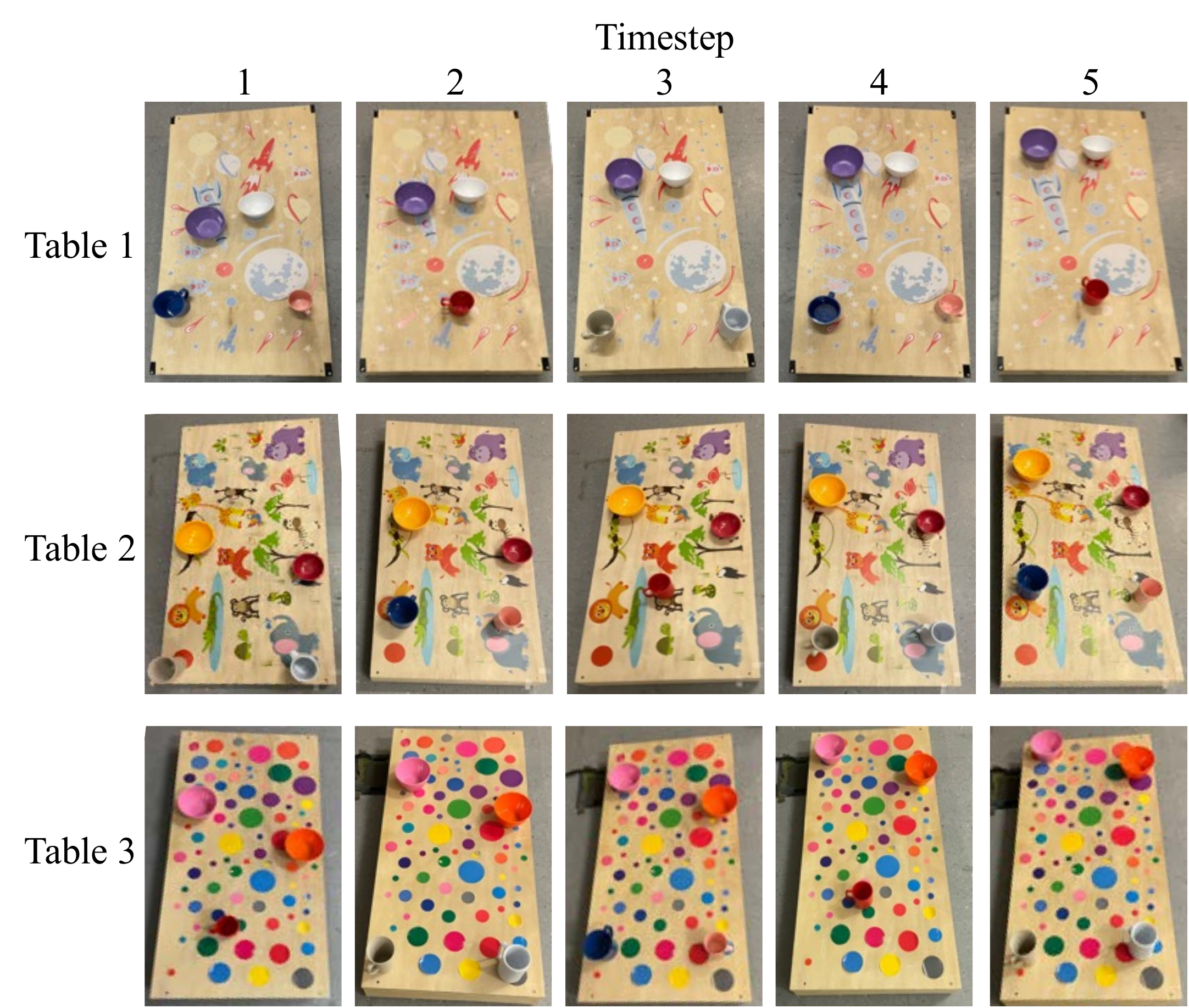}
\caption{\small \textbf{Real Object Dynamics.} Illustration of object dynamics across tables across each of 5 different timesteps. Bowls displace vertically on a table while mugs teleport between adjacent tables.}
\label{fig:object_movement}
\vspace{-5pt}
\end{figure}

%% file: new_fig_iros/data_preprocess.tex
\begin{figure}
\centering
\includegraphics[width=\linewidth]{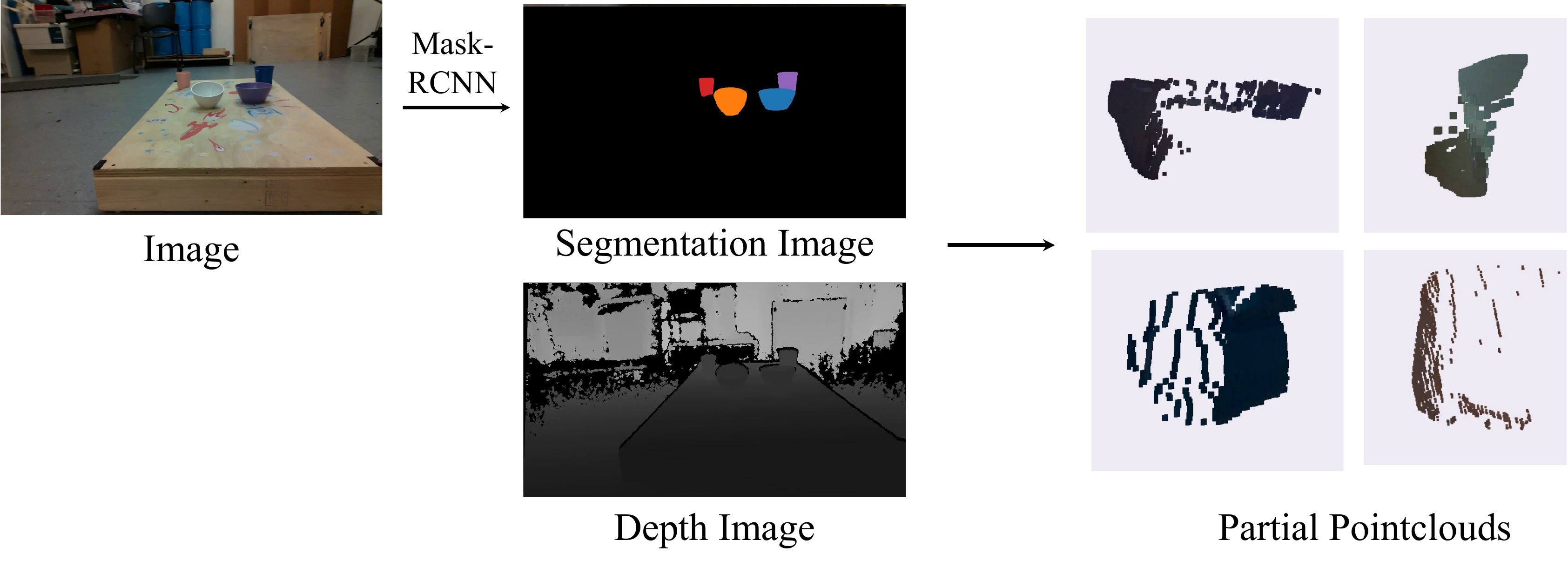}
\vspace{-15pt}
\caption{\small \textbf{Data Processing.}  An input image is fed through a Mask-RCNN model to obtain a semantic segmentation of bowls and mugs. The resultant segmentation is projected with the observed depth image to obtain partial pointclouds of objects.}
\label{fig:data_preprocess}
\vspace{-20pt}
\end{figure}

%% file: new_fig_iros/3d_reconstruct_real_side.tex
\begin{figure}[t]
\centering
\includegraphics[width=1.0\linewidth]{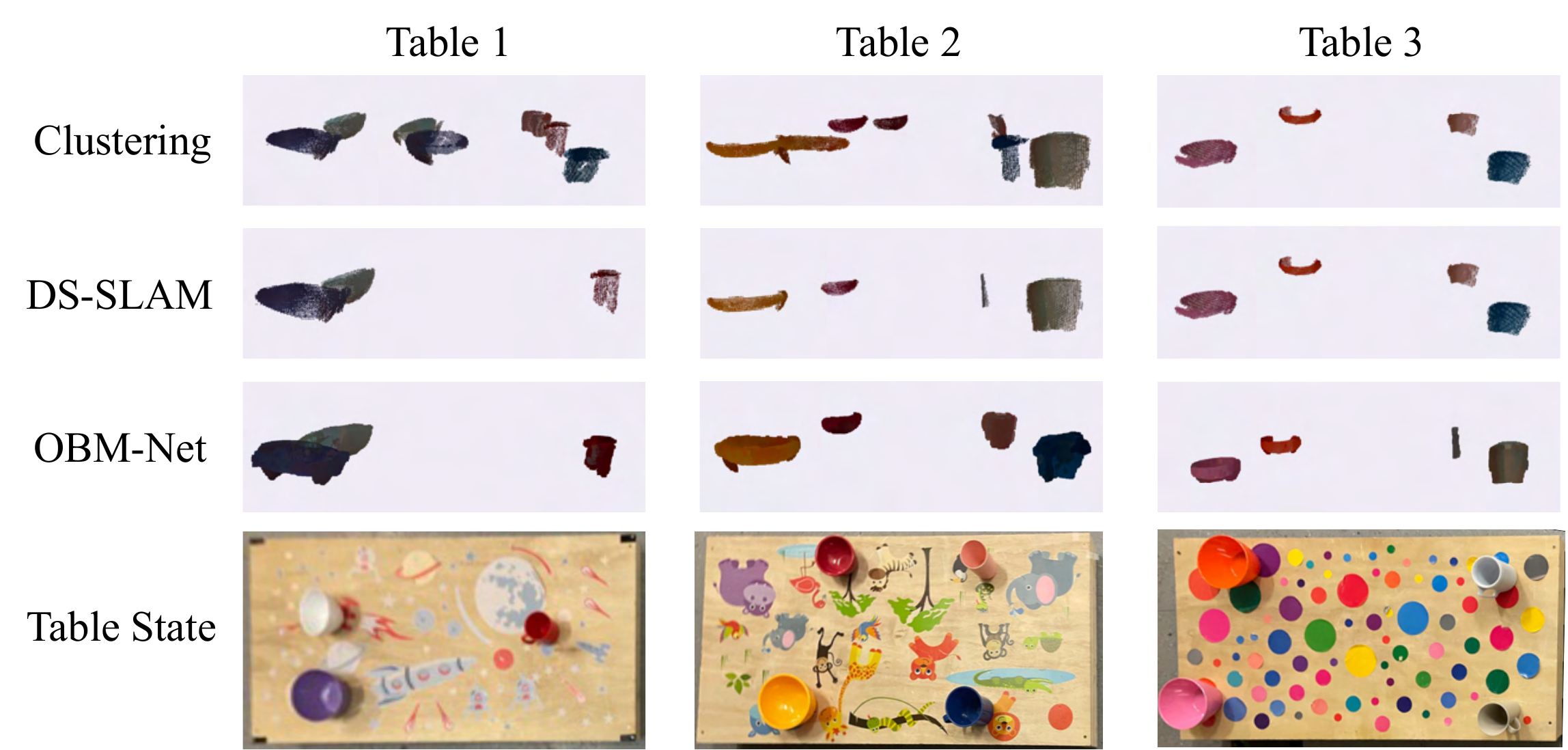}
\caption{\small \textbf{Real 3D Shape Reconstructions (side).} Illustration of a side view of reconstructed objects from \model compared to baselines. \model is able to more accurately reconstruct the shape of objects, aggregating past observations at both the current table and other tables. }
\label{fig:3d_reconstruct_side}
\vspace{-20pt}
\end{figure}

%% file: new_fig_iros/real_robot_eval.tex
\begin{table}[t]
\footnotesize
\setlength{\tabcolsep}{3.5pt}
\centering
\resizebox{\linewidth}{!}{
\begin{tabular}{@{}lcccccc@{}}
    \toprule
     &  \multicolumn{2}{c}{Table 1} & \multicolumn{2}{c}{Table 2} & \multicolumn{2}{c}{Table 3} \\
      \cmidrule(lr){2-3} \cmidrule(lr){4-5} \cmidrule(lr){6-7} 
     & Position & Object &  Position & Object & Position & Object \\
     & Error & Accuracy &  Error & Accuracy & Error & Accuracy\\
    \midrule
    Clustering &  0.198 & 67\% & 0.181 & 50\% & 0.119 & 50\% \\
    DS-SLAM~\cite{yu2018ds} & 0.028 & 100\% & 0.096 & 50\% & 0.119 & 50\% \\
    \model & \textbf{0.025} & \textbf{100\%} & \textbf{0.035} & \textbf{100\%} & \textbf{0.035} & \textbf{100\%}   \\
    \bottomrule
\end{tabular}
}
\caption{\small \textbf{Quantitative Analysis of \model on Real Data.} Quantitative comparison of \model with baselines on regressing the final positions and presence of individual objects at each table at the final timestep 5. \model outperforms baselines.
}
\label{tbl:comparison_robotics_real}
\vspace{-10pt}
\end{table}

%% file: new_fig_iros/3d_reconstruct_real_top.tex
\begin{figure}[t]
\centering
\includegraphics[width=1.0\linewidth]{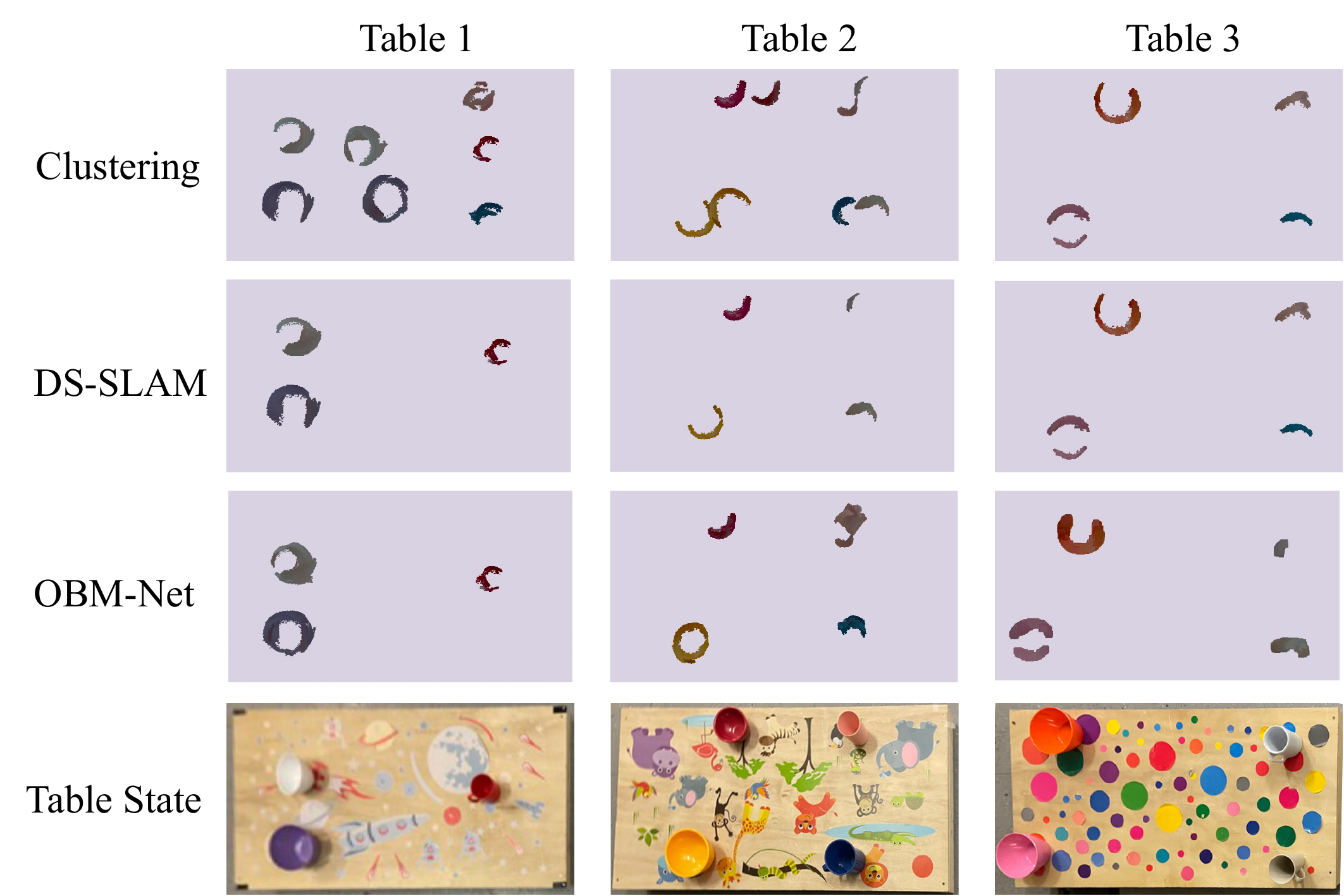}
\caption{\small \textbf{Real 3D Shape Reconstructions (top).} Illustration of top view of reconstructed object shapes from \model compared to each baselines. Compared to baselines, \model is able to simulate the horizontal movement of bowls, as well as the dynamic movement of mugs. Reconstructions of individual objects, such as mugs in table 2 is aided by observations at other tables.}
\label{fig:3d_reconstruct_top}
\vspace{-20pt}
\end{figure}

%% file: text/discussion.tex
\section{Discussion}
This work has demonstrated an approach to constructing a dynamic object-based memory for household robots in changing environments.  By incorporating algorithmic bias inspired by a classical solution to the problem of filtering to estimate the state of multiple objects simultaneously, coupled with modern machine-learning techniques,  \model learns from experience {\em how to perform as a data-association filter} in novel environments with complex observation and transition models that would be too difficult to hand-specify.  Importantly, our system may be applied with {\em no prior knowledge} about the types of observations or desired output hypotheses or the frequency of observations.  We have demonstrated its effectiveness in a variety of problems including simulated and real robot object memory problems.